\documentclass[lettersize,journal]{IEEEtran}
\usepackage{amsmath,amsfonts}
\usepackage{algorithmic}
\usepackage{algorithm}
\usepackage{array}
\usepackage[caption=false,font=normalsize,labelfont=sf,textfont=sf]{subfig}
\usepackage{textcomp}
\usepackage{stfloats}
\usepackage{url}
\usepackage{verbatim}
\usepackage{graphicx}
\usepackage{titlesec} 
\usepackage{cite}
\usepackage{float}
\usepackage{subcaption}
\usepackage{multirow}
\usepackage{multicol}
\usepackage{tablefootnote}
\usepackage{booktabs}
\begin{document}

\title{FMANet: A Novel Dual-Phase Optical Flow Approach with Fusion Motion Attention Network for Robust Micro-expression Recognition}

\author{%
    Luu Tu Nguyen\IEEEauthorrefmark{1}, 
    Vu Tram Anh Khuong\IEEEauthorrefmark{1},
    Thi Bich Phuong Man\IEEEauthorrefmark{1},
    Thi Duyen Ngo\IEEEauthorrefmark{1}, 
    Thanh Ha Le\IEEEauthorrefmark{1}\thanks{Corresponding author: Thanh Ha Le (ltha@vnu.edu.vn)}\\
    \IEEEauthorblockA{\IEEEauthorrefmark{1}Faculty of Information Technology, VNU University of Engineering and Technology, Ha Noi, Viet Nam}\\
    % \IEEEauthorblockA{\IEEEauthorrefmark{2}Department of Electrical Engineering, Institute Name, City, Country}\\
    % \IEEEauthorblockA{\IEEEauthorrefmark{3}Department of Artificial Intelligence, Organization Name, City, Country}\\
    % Email: first.author@email.com, second.author@email.com, third.author@email.com, fourth.author@email.com
}

\markboth{IEEE Transactions on Affective Computing,~Vol.~14, No.~8, December~2024}%
{Shell \MakeLowercase{\textit{et al.}}: A Sample Article Using IEEEtran.cls for IEEE Journals}

% \IEEEpubid{2024 IEEE}

\maketitle

\begin{abstract}
Facial micro-expressions, characterized by their subtle and brief nature, are valuable indicators of genuine emotions. Despite their significance in psychology, security, and behavioral analysis, micro-expression recognition remains challenging due to the difficulty of capturing subtle facial movements. Optical flow has been widely employed as an input modality for this task due to its effectiveness. However, most existing methods compute optical flow only between the onset and apex frames, thereby overlooking essential motion information in the apex-to-offset phase.
To address this limitation, we first introduce a comprehensive motion representation, termed Magnitude-Modulated Combined Optical Flow (MM-COF), which integrates motion dynamics from both micro-expression phases into a unified descriptor suitable for direct use in recognition networks. Building upon this principle, we then propose FMANet, a novel end-to-end neural network architecture that internalizes the dual-phase analysis and magnitude modulation into learnable modules. This allows the network to adaptively fuse motion cues and focus on salient facial regions for classification. Experimental evaluations on the CASME-II, SAMM and MMEW datasets, widely recognized as standard benchmarks, demonstrate that our proposed MM-COF representation and FMANet outperforms existing methods, underscoring the potential of a learnable, dual-phase framework in advancing micro-expression recognition.
\end{abstract}


\section{Introduction}
\label{s: introduction}
\IEEEPARstart{F}{acial} expressions play a crucial role in conveying emotions and facilitating communication. Expressions are generally categorized into two main types: one type, known as macro-expressions, is characterized by strong intensity and longer duration \cite{ekman1997face}, while the other type, known as micro-expressions (ME), is more challenging to detect due to its extremely brief occurrence, typically lasting less than 0.5 seconds \cite{shen2012effects}. ME consist of subtle, involuntary movements with low intensity. Owing to their unique characteristics, micro-expressions have increasingly attracted attention in research and practical applications, as they reveal genuine emotions that individuals may attempt to conceal. They play a significant role in fields such as psychology \cite{Bhushan2015}, security \cite{yan2013fast}, and behavioral analysis \cite{polikovsky2010detection}. Due to their transient and subtle nature, accurately recognizing micro-expressions remains a significant challenge, requiring sophisticated analytical techniques. To address this challenge, researchers have explored various micro-expression recognition (MER) methods, among which deep learning has emerged as the most effective approach in recent years. Deep learning offers powerful feature extraction capabilities, allowing models to learn complex spatial and temporal patterns directly from ME data. Two critical factors directly influencing the performance of MER using deep learning are the input modality and the MER network architecture.

Input modality in MER can be broadly categorized into appearance-based and motion-based approaches. Appearance-based methods rely on static features extracted from individual frames, such as texture, shape, and facial landmarks. While these methods can capture spatial information, they often struggle to effectively detect the fleeting and low-intensity movements that define micro-expressions. In contrast, motion-based approaches analyze facial movement patterns directly, making them more effective in recognizing micro-expressions \cite{review1}. Among motion-based methods, optical flow-based techniques have gained prominence due to its ability to capture subtle facial movements while minimizing identity-specific features \cite{article55, xia2019spatiotemporalrecurrentconvolutionalnetworks}. Various studies have explored different ways to leverage optical flow for micro-expression analysis, each with its own strengths and limitations. Early approaches such as Bi-WOOF \cite{biwoof} and MDMO \cite{MDMO} primarily relied on handcrafted features extracted from optical flow. Despite their effectiveness in highlighting key motion patterns, these methods suffered from limited representational capacity and were highly sensitive to noise in low-texture regions. Unlike the aforementioned works which exploited only the single dominant direction of optical flow in each facial region, Allaert et al. \cite{RHPM} introduced a refined motion representation by emphasizing coherent optical flow structures. However, these approaches often involved complex architectures, increasing computational costs and requiring large-scale datasets for effective training.
% However, by focusing solely on the peak intensity frame, OFF-ApexNet may overlook valuable motion cues from the onset and offset frames.

Despite significant advancements, most existing MER methods utilizing optical flow primarily focus on the transition from the onset frame (the initial frame of the micro-expression) to the apex frame (the peak intensity frame), capturing only the buildup phase of motion. However, this approach overlooks the apex-to-offset phase (the transition from the peak to the final frame), which is crucial for understanding the resolution of the expression \cite{review1, review2}. While Liu et al. \cite{9336412} incorporated the offset frame in their work, its contribution was limited. They employed a five-stream CNN where the offset-related motion was just one of five inputs, and the authors failed to clearly articulate its specific impact on the network's performance. Moreover, their model's accuracy did not show a clear advantage despite using multiple input types. This common omission of the apex-to-offset phase results in an incomplete temporal representation, hindering recognition accuracy. 
% Additionally, optical flow-based methods face challenges in handling the inherently subtle and brief nature of micro-expressions. They are particularly susceptible to noise from irrelevant facial regions or areas with low texture contrast, which can degrade motion estimation accuracy. The absence of the apex-to-offset transition further limits the completeness of micro-expression representation, ultimately affecting classification performance.
To address these limitations, a more comprehensive approach is needed, one that not only models all phases of a micro-expression but also enhances the motion representation to be more robust against noise.

In addition to input modality, the MER network architecture plays a crucial role in optimizing performance. Recent advances in deep learning (e.g., convolutional neural network (CNN)), have significantly improved MER performance \cite{review1}. Deep CNN have been widely used in MER to capture both spatial and temporal patterns of micro-expressions. Spatial CNN, such as ResNet \cite{Lai2023}, VGG \cite{9567872} and Inception \cite{Zhou2023}, extract fine-grained facial features from static frames, particularly the apex frame. However, they fail to capture temporal motion. Temporal networks such as LSTM \cite{10.1007/978-3-031-27066-6_2}, GRU \cite{ inproceedings}, and 3D-CNN \cite{10674479} analyze frame sequences to model motion patterns over time. 
More advanced models such as 3D-FCNN \cite{3d}, CNNCapsNet \cite{CNNCapsNet}, and MSCNN \cite{MSCNN} incorporated optical flow into deep spatial-temporal frameworks. While effective, these models require large datasets and are computationally expensive. Hybrid spatial-temporal models such as Inception-LSTM \cite{9389913} and 3D-ResNet \cite{Gajjala_2021} integrate both spatial and temporal learning but further increase complexity. Although Deep CNN deliver strong performance in many domains, their reliance on large datasets makes them prone to overfitting in MER, where data is often limited. To address the data limitations of MER, researchers have explored shallow CNN, achieving strong performance with reduced complexity \cite{gan2018bi, merastc, compound, off-apex}.
% Several studies highlight the advantages of shallow networks.  Improved MobileNetV2 \cite{mobile} and OF-PCANet \cite{pcanet} further explored lightweight deep architectures for optical flow feature extraction, but their reliance on pre-defined motion representations limited their adaptability to varying micro-expression patterns. Belaiche et al. \cite{belaiche2020cost} constructed a shallow network by reducing the depth of ResNet through the removal of multiple convolutional layers. Peng et al. \cite{macro} introduced ResNet10, a lightweight variant of ResNet, for more efficient feature representation. 
% % Zhao et al. \cite{compound} proposed a 6-layer CNN incorporating a 1×1 convolutional layer after the input layer to enhance non-linear feature extraction without increasing computational complexity.
% \begin{figure*}[h]
%     \centering
%     \includegraphics[width=0.9\linewidth]{figures/Untitled Diagram.drawio (10).png}
%     \caption{The workflow of the MER method with Magnitude-Modulated Combined Optical Flow}
%     \label{fig:pipline}
% \end{figure*}
% Takalkar et al. \cite{dual} developed a compact architecture consisting of five convolutional layers, three max-pooling layers, and three fully connected layers. These studies demonstrate that Shallow CNN not only mitigate overfitting but also outperform Deep CNN in MER when data is limited. 
However, despite these advancements, existing approaches still face challenges in effectively learning expressive emotion representations under limited micro-expression data conditions. While some methods focus primarily on reducing model size, potentially at the expense of representational capacity, others introduce modifications that do not always lead to significant improvements in recognition performance.

Building upon the limitations remained in previous MER approaches, this paper proposes a novel approach that integrates both phases of a micro-expression (onset-to-apex and apex-to-offset) into a MM-COF representation. This representation effectively combines the optical flows from both phases while modulating their magnitudes to emphasize critical regions and suppress noisy ones, resulting in a comprehensive depiction of micro-expressions. Furthermore a Shallow Convolutional Neural Network has been proposed for the classification of these enhanced features. Furthermore, we proposed Fusion Motion Attention Network (FMANet), a novel neural network architecture that internalizes the principles of dual-phase motion analysis and magnitude modulation into learnable modules. At the core of our architecture are two innovative components: a Phase-Aware Consensus Fusion Block (FFB), which replaces handcrafted fusion rules with a data-driven mechanism to adaptively integrate feature maps from both motion phases based on a learned consensus. And a Soft Motion Attention Block (SMAB), which reformulates hard-thresholding into a differentiable attention mechanism to selectively amplify salient motion features.
% This approach not only ensures the full utilization of motion data but also optimizes the input to achieve the best possible performance in micro-expression recognition.

The contributions of this paper include a comprehensive approach to MER, which consists of three key components:
\begin{itemize}
    % \item We emphasize the importance of full-phase micro-expression analysis by incorporating apex-offset frames in addition to the commonly used onset-apex frames. Unlike most previous studies that focus only on the onset-apex phase, we demonstrate that analyzing all three key frames (onset, apex, and offset) significantly improves micro-expression recognition (MER) performance. This comprehensive approach ensures that no critical temporal information is discarded.  

    % \item We propose a novel method, Combined Optical Flow (COF), which integrates two optical flow representations: onset-apex and apex-offset (see section \ref{s: proposed method}). By leveraging complementary motion patterns from both segments, our approach captures subtle expression transitions more effectively than conventional single-phase methods. This contribution addresses a major limitation in prior research, where the offset frame is often neglected despite its potential to enhance MER performance.  

    %  \item We extend the idea of COF and propose the Magnitude-Modulated Combined Optical Flow (MM-COF), which enhances high-intensity motion regions while suppressing low-intensity areas. This modulation improves feature discriminability, enabling MM-COF to more effectively capture and differentiate micro-expressions with varying motion intensities.
    
    \item A motion representation called Magnitude-Modulated Combined Optical Flow is proposed for micro-expression recognition (see Section \ref{s: proposed method}).

    \item A shallow convolutional neural network has been proposed specifically for MER using MM-COF input modality (see Section \ref{s: proposed method}).

    \item A novel architecture Fusion Motion Attention Network, a network that embeds the principles of MM-COF into learnable modules (see Section III).
\end{itemize}

The proposed method has been evaluated by comparing its accuracy with state-of-the-art approaches (see Section \ref{s: experiment and results}). The experimental results demonstrate the effectiveness of MM-COF representation  with Shallow CNN particularly in handling imbalanced micro-expression datasets and FMANet in the MER task. Furthermore, an comprehensive ablation study has been conducted on each contribution of the paper, ensuring that each individual element meaningfully contributes to the overall advancements in modern MER research.

\begin{figure*}
    \centering
    \includegraphics[width=0.9\linewidth]{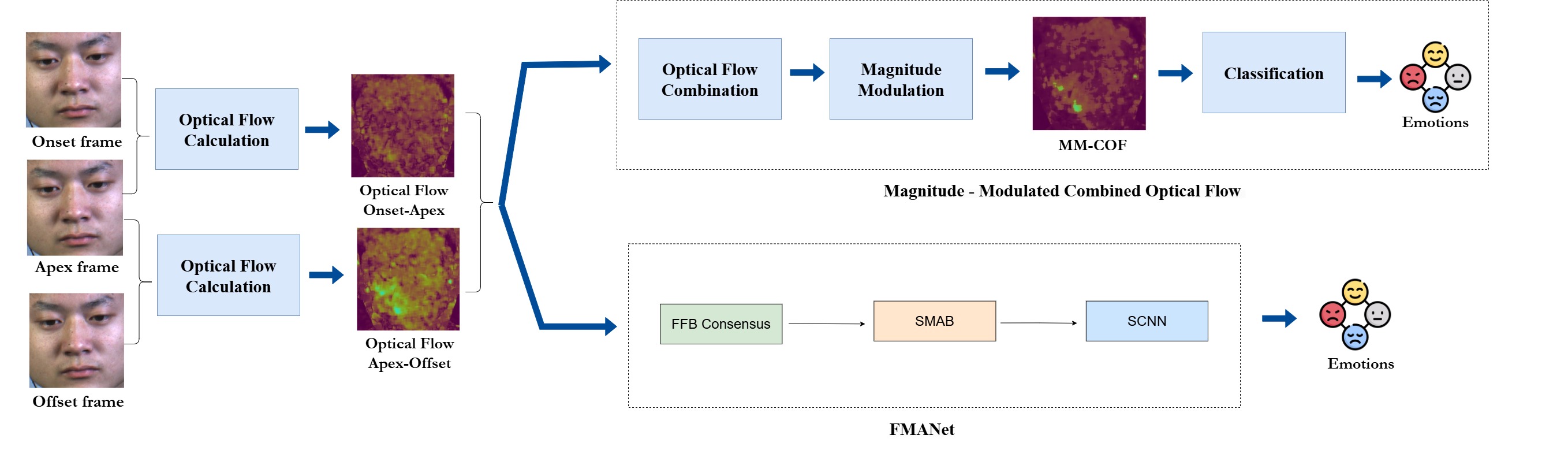}
    \caption{The workflow of our proposed method}
    \label{fig:pipeline}
\end{figure*}

\section{Magnitude-Modulated Combined Optical Flow Representation}
\label{s: proposed method}
% Micro-expressions unfold through asymmetric temporal phases: the onset–apex phase captures the emergence and peak intensity, while the apex–offset phase reflects relaxation and transition. Assuming symmetry between these phases oversimplifies the dynamics and risks discarding informative offset cues. This challenge is exacerbated by the fact that apex frames are rarely centered between onset and offset in existing MER datasets.
Micro-expressions unfold across two distinct temporal phases: an onset-to-apex (formation) phase and an apex-to-offset (relaxation) phase. While the complete motion in these phases is not perfectly identical, exhibiting asymmetry due to non-key motions, subtle variations in muscle relaxation, or temporal shifts, we posit that the core muscle activations defining the genuine emotion are fundamentally symmetric. Therefore, our fusion of the two stages is motivated by the goal of filtering out the asymmetric non-key motions while retaining and reinforcing only the symmetric key motions. This approach allows us to distill a more robust and purified representation of the true expression, a challenge exacerbated by the fact that apex frames are rarely centered in existing MER datasets.

Figure~\ref{fig:pipeline} illustrates the overall workflow of our proposed framework. Given an onset, apex, and offset frame, optical flow is first calculated to capture motion between onset-apex and apex-offset phases. These dual-phase flows form the basis for two complementary pipelines. The upper branch corresponds to the MM-COF representation, where onset–apex and apex–offset flows are combined, modulated by motion magnitude, and then classified using a lightweight CNN. This highlights our contribution in designing a more discriminative optical flow representation. The lower branch shows the end-to-end FMANet model, which extends the MM-COF idea into a dual-stream network. It integrates both motion phases through the FFB consensus module and SMAB attention block, followed by SCNN for classification. This design enables joint learning of complementary dynamics within a unified architecture.
% To capture the full temporal evolution, we first construct the MM-COF, which fuses onset–apex and apex–offset flows and applies magnitude-based modulation to highlight salient motion while suppressing noise. A compact Shallow CNN is then employed to classify the resulting representation, mitigating overfitting on limited data.

% The overall workflow of MM-COF and its end-to-end extension is illustrated in Fig.~\ref{fig:pipline}, and the following subsections detail each component.

% Lược phần  Optical Flow Calculation cho bớt lan man :) gộp OF calc + combine

\subsection{Optical Flow Calculation}
\label{s: optical flow calculating}

Optical flow is a method used to estimate the motion of objects between consecutive frames in a video sequence. It is commonly applied in motion analysis tasks and is subject to certain conditions to ensure its effectiveness. 

% To estimate the optical flow, it is generally assumed that: 
% \begin{itemize}
%     \item The brightness of the moving objects remains unchanged between the source and target frames. 
%     % Thus the noises generated by a large variety of imaging variables such as the shadows, highlights, illumination and surface translucency phenomena are entirely neglected.
%     \item The displacement of each pixel between consecutive frames is minimal, as object motion occurs gradually over time.
%     % \item Image flow field is continuous and differentiable in both the space and time domains.
%     \item The background and objects belonging to it are assumed to be static.
% \end{itemize}

The onset, apex, and offset frames serve as critical anchor points that capture the entire life-cycle of a micro-expression: its formation, peak intensity, and subsequent relaxation. Therefore the optical flow from the two phases will be calculated by frame onset and apex, frame apex and offset.

Once optical flow is computed, it generates a dense motion field in which each pixel  \((x,y)\) is assigned a motion vector \((u(x, y), v(x, y))\), where \((u(x, y)\) and \(v(x, y))\) represent the horizontal and vertical components of movement, respectively. The magnitude \( M(x, y) \) of this motion vector measures the motion intensity at each pixel and serves as a crucial feature for analyzing facial dynamics, as shown in Equation \ref{eq:magnitude}.

\begin{equation}
\label{eq:magnitude}
M(x, y) = \sqrt{u(x, y)^2 + v(x, y)^2}
\end{equation}

This magnitude serves as a critical feature for analyzing the dynamics of facial expressions, as it quantifies the intensity of motion for each pixel. 

% Micro-expressions evolve through three temporal stages: onset, apex, and offset. To capture their subtle dynamics, we compute optical flow between the onset-to-apex and apex-to-offset segments, which respectively characterize the rising and relaxing phases of the expression. Optical flow is estimated using the Farneback algorithm \cite{farneback}, which models the neighborhood of each pixel with a quadratic polynomial and derives displacement fields by matching polynomial coefficients across consecutive frames. This provides robust pixel-wise motion estimation, well-suited for subtle facial movements.  

\subsection{Optical Flow Combination}
\label{s: Combined Optical Flow}

% Micro-expressions evolve through three temporal stages: onset, apex, and offset. To capture their subtle dynamics, we compute optical flow between the onset-to-apex and apex-to-offset segments, which respectively characterize the rising and relaxing phases of the expression. Optical flow is estimated using the Farneback algorithm \cite{farneback}, which models the neighborhood of each pixel with a quadratic polynomial and derives displacement fields by matching polynomial coefficients across consecutive frames. This provides robust pixel-wise motion estimation, well-suited for subtle facial movements.  

Empirical observations indicate that action units activated in the onset–apex phase often reappear in the apex–offset phase, producing consistent motion patterns across both segments. Consequently, regions with strong expression-related motion exhibit high intensity in both optical flow maps. Motivated by this property, we combine the two flows into a unified representation using a weighted summation, where each phase contributes proportionally to the overall micro-expression motion.  

To ensure scale invariance, the flow magnitudes of each phase \((M_1(x,y), M_2(x,y))\) are normalized as follows:

\begin{equation}
\label{eq:normalize1}
M_{1}^{norm}(x,y) = \frac{M_1(x,y) - \min(M_1)}{\max(M_1) - \min(M_1)}
\end{equation}

\begin{equation}
\label{eq:normalize2}
M_{2}^{norm}(x,y) = \frac{M_2(x,y) - \min(M_2)}{\max(M_2) - \min(M_2)}
\end{equation}

After normalization, the two optical flow representations are combined by computing a weighted sum at each spatial location, as expressed in the Equation \ref{eq:fusion}.

\begin{equation}
\label{eq:fusion}
M_{c}(x,y) = \theta_1 M_{1}^{norm}(x,y) + \theta_2 M_{2}^{norm}(x,y)
\end{equation}
where \(M_c(x,y)\) represents the combined optical flow magnitude at each spatial location \((x,y)\), \(\theta_1\)  and \(\theta_2\)  are weighting coefficients representing the relative importance of each phase. The selection of \(\theta_1, \theta_2\) is based on the assumed contribution of each phase to the overall micro-expression process. For instance, if both phases are considered equally important, the weights can be set as \(\theta_1 = \theta_2 = 1\). The choice of \(\theta\) values directly influences the effectiveness of MER, as different weight configurations may impact motion representation quality. A detailed analysis of its impact on MER performance is provided in the following section.

Fig.~\ref{fig:histogram}b shows the result of the proposed optical flow combination method, referred to as Combined Optical Flow (COF). This synthesized optical flow representation emphasizes consistent motions from both phases, accurately reflecting the motion dynamics of micro-expressions. It provides continuous information on motion intensity throughout the entire micro-expression process, which improves the accuracy of micro-expression recognition.

\subsection{Magnitude Modulation} \label{s: Combined Optical Flow with threshold}
% While the combined optical flow described in Section \ref{s: Combined Optical Flow} provides a unified representation of motion dynamics, challenges may arise due to the presence of irrelevant motion or noise, as well as the varying significance of different regions. To address these issues, the Magnitude-Modulated Combined Optical Flow (MM-COF) method is proposed. MM-COF incorporates a thresholding mechanism to selectively enhance significant motion regions while suppressing less relevant or noisy areas. The complete process of the MM-COF method is detailed in Algorithm \ref{alg:COF}.

% The thresholding approach is designed to refine the optical flow representation by prioritizing regions with higher motion intensity, which are more likely to correspond to critical micro-expression features. By filtering out low-intensity motions that may introduce noise or redundancy, MM-COF ensures that the final representation retains only the most pertinent motion information. This refinement is crucial for improving the accuracy and robustness of subsequent micro-expression recognition tasks. 

% \subsubsection{\textbf{Thresholding mechanism}}
% \label{s: Thresholding Mechanism}

\begin{figure}
    \centering
    \includegraphics[width=0.7\linewidth]{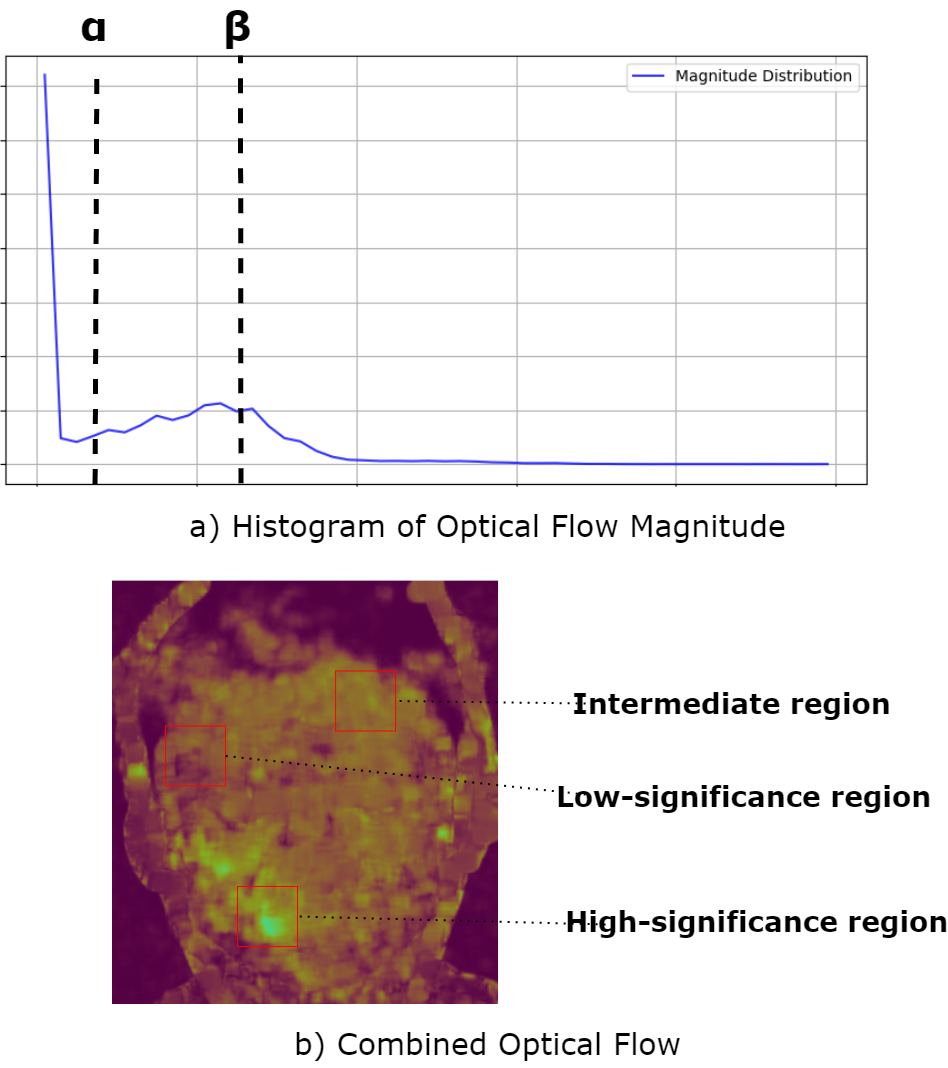}
    \caption{Visualization of Optical Flow Magnitude in Combined Optical Flow.}
    \label{fig:histogram}
\end{figure}

Although the Combined Optical Flow (COF) integrates motion from both phases into a single map, its ability to represent subtle facial changes is limited by noise and irrelevant movements. To refine this representation, we propose a magnitude modulation strategy that adjusts motion intensity according to its significance, suppressing noise while emphasizing discriminative regions.  

Fig.~\ref{fig:histogram}a shows the distribution of COF magnitudes $M_c$, which can be divided into three regions corresponding to different levels of motion significance (Fig.~\ref{fig:histogram}b). Low-magnitude regions often reflect noise or inconsequential movements, intermediate regions capture subtle but relevant facial dynamics, and high-magnitude regions contain the most discriminative motion patterns. To model these differences, two thresholds \(\alpha\) and \(\beta\) \((\alpha < \beta)\) are introduced, partitioning the magnitude space into:  

% ensuring that the most relevant information is emphasized while minimizing the impact of noise:
\begin{itemize}
    \item \textbf{Low-significance region} $(M_c < \alpha)$: 
    This region captures minor and often inconsequential motion, such as noise or irrelevant movements that do not contribute significantly to the expression's dynamics. 
    % Examples include slight random movements or background noise.
    % To reduce their impact, the magnitude of these motions is adjusted using a weight \(w_1\), ensuring that the recognition network focuses on more relevant features while minimizing noise.

    \item \textbf{Intermediate region} $(\alpha \leq M_c \leq \beta)$: 
    This region represents moderate motion patterns that are relevant but less critical than the high-significance region. These movements reflect subtler facial dynamics that contribute to the overall expression but do not capture its peak intensity. 
    % Motion in this region is preserved without modification, as it still provides valuable information for micro-expression recognition.

    \item \textbf{High-significance region} $(M_c > \beta)$: 
    This region contains the most significant motion patterns, often associated with pronounced facial movements during a micro-expression's peak intensity. These patterns are critical for distinguishing expressions.  
    % To highlight their importance, the magnitudes in this region are adjusted using a weight \(w_2\), enhancing key dynamics and making them more prominent in the final representation.
\end{itemize}

To further refine the optical flow representation and ensure that the most relevant information is emphasized while minimizing the impact of noise, a magnitude modulation approach is proposed in this paper. This method adjusts motion intensities within each region, enhancing critical features while suppressing irrelevant variations. The modulation process is defined by the Equation \ref{eq: eq8}.
\begin{equation}
\label{eq: eq8}
M_{mod}(x,y) =
\begin{cases}
w_1 \cdot M_c(x,y) & \text{if } M_c(x,y) > \beta \\
w_2 \cdot M_c(x,y) & \text{if } M_c(x,y) < \alpha \\
M_c(x,y) & \text{otherwise}
\end{cases}
\end{equation}
where \(M_{mod}(x,y)\) represents the combined optical flow magnitude modulated at each spatial location \((x,y)\), $w_1$ and $w_2$ are weighting factors that regulate the emphasis on high-significance regions and the attenuation of low-significance regions, respectively. 

To achieve optimal segmentation, two distinct approaches to threshold selection are considered: manual thresholding and adaptive thresholding. Each method offers unique advantages and challenges, with their suitability depending on the specific requirements of the micro-expression recognition system.

\begin{figure}[h]
    \centering
    \includegraphics[width=1\linewidth]{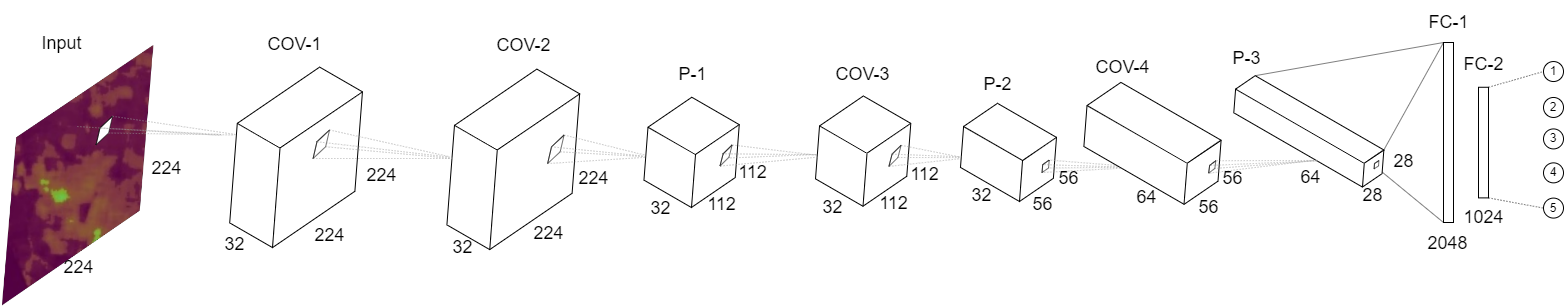}
    \caption{Our proposed Shallow Convolutional Neural Network (SCNN) architecture. The input data is MM-COF images. It is then processed by four convolutional layers and three pooling layers, followed by two fully connected layers.}
    \label{fig:SCNN}
\end{figure}
\subsubsection{Manual thresholding}
This involves setting the values of $\alpha$ and $\beta$ based on empirical observations or prior experimental calibration. In this approach, the thresholds are predetermined, often through trial and error or established benchmarks, to effectively differentiate between insignificant noise and relevant facial motion. While this method can yield effective results, it requires expertise and may not generalize well across varying datasets or facial expression characteristics. 

\subsubsection{Adaptive thresholding} 
To improve robustness of manual threshold selection, we propose a data-driven adaptive thresholding mechanism that dynamically adjusts \(\alpha\) and \(\beta\) based on the statistical distribution of motion magnitudes. Given the combined magnitude map \(M_{c}(x,y)\), the adaptive thresholds are computed in Equation \ref{eq:eq7}.

\begin{equation}
\label{eq:eq7}
\begin{aligned}
\mu = \frac{1}{N} \sum_{x,y} M_c(x,y) \\
\sigma = \sqrt{\frac{1}{N} \sum_{x,y} (M_c(x,y) - \mu)^2} \\
\alpha_{adaptive} = \mu + k_{upper} \cdot \sigma \\
\beta_{adaptive} = \mu - k_{lower} \cdot \sigma
\end{aligned}
\end{equation}

where \(\alpha_{adaptive}\) and \(\beta_{adaptive}\) are adaptive parameters for \(\alpha\) and \(\beta\), which are dynamically selected based on each individual data sample. \(\mu\) and \(\sigma\) represent the mean and standard deviation of \(M_{C}\), \(N\) is the total number of pixels, and \(k_{upper}, k_{lower}\) are tunable coefficients controlling threshold strictness. In our implementation, \(k_{upper} = 2\) and \(k_{lower} = 1\) empirically provided optimal noise suppression while preserving micro-expression-related motions.

By applying this modulation, the COF is transformed into the MM-COF, in which irrelevant motion is attenuated and salient facial dynamics are amplified. This yields a refined representation that enhances feature contrast and improves robustness to noise, providing a more discriminative input for subsequent recognition.

\subsection{Classification}
\label{s: classification}

The Magnitude-Modulated Combined Optical Flow (MM-COF), described in Section \ref{s: Combined Optical Flow with threshold}, serves as the input representation for micro-expression recognition. To address the performance limitations of deep convolutional neural network (DCNN) in MER, particularly in data-limited scenarios, this study proposes a shallow convolutional neural network (SCNN) specifically optimized for MER. The SCNN architecture is designed with fewer layers and smaller convolutional kernel sizes, enabling efficient feature extraction and classification. Fig. \ref{fig:SCNN} illustrates the conceptual visualization of our proposed SCNN architecture. The SCNN model consists of four convolutional layers, three max-pooling layers, and two fully connected layers. ReLU activation functions are applied after each convolutional and fully connected layer to enhance nonlinearity and improve the model’s feature learning capability. The detailed architecture is described in Table~\ref{tab:mycnn_layers}.

\begin{table}[h]
\centering
\caption{SCNN configuration with four convolution layers, three pooling layers, two fully connected layers, and an output layer}
\resizebox{\linewidth}{!}{ % Điều chỉnh bảng để vừa với trang
\begin{tabular}{|c|c|c|c|c|c|}
\hline
\textbf{Layer} & \textbf{Filter Size} & \textbf{Kernel} & \textbf{Stride} & \textbf{Padding} & \textbf{Output Size} \\ 
\hline
Conv 1  & $3 \times 3 \times 32$  & $3 \times 3$  & 1  & 1  & $32 \times 224 \times 224$ \\ 
\hline
Conv 2  & $3 \times 3 \times 32$  & $3 \times 3$  & 1  & 1  & $32 \times 224 \times 224$ \\ 
\hline
Pool 1  & -  & $2 \times 2$  & 2  & 0  & $32 \times 112 \times 112$ \\ 
\hline
Conv 3  & $3 \times 3 \times 32$  & $3 \times 3$  & 1  & 1  & $32 \times 112 \times 112$ \\ 
\hline
Pool 2  & -  & $2 \times 2$  & 2  & 0  & $32 \times 56 \times 56$ \\ 
\hline
Conv 4  & $3 \times 3 \times 32$  & $3 \times 3$  & 1  & 1  & $64 \times 56 \times 56$ \\ 
\hline
Pool 3  & -  & $2 \times 2$  & 2  & 0  & $64 \times 28 \times 28$ \\ 
\hline
FC 1  & -  & -  & -  & -  & $1024 \times 2$ \\ 
\hline
FC 2  & -  & -  & -  & -  & $1024 \times 1$ \\ 
\hline
Output  & -  & -  & -  & -  & $5 \times 1$ \\ 
\hline
\end{tabular}
}
\label{tab:mycnn_layers}
\end{table}

Our proposed SCNN is specifically designed to address the challenges of micro-expression recognition by focusing on subtle and localized motion patterns. The shallow architecture, combined with small kernel sizes and reduced spatial dimensions, enhances the model's ability to capture fine-grained features critical for MER. Additionally, the use of magnitude-modulated combined optical flow as input improves the model's capacity to capture temporal dynamics, leading to better recognition performance. This design aims to improve accuracy and adaptability in MER tasks, particularly in scenarios with limited data.

Although effective, MM-COF depends on fixed fusion weights and manually tuned thresholds, limiting adaptability across subjects and datasets. To overcome this, the next section introduces FMANet, an end-to-end extension that embeds the principles of MM-COF into learnable modules. 

\section{Fusion Motion Attention Network}
\label{s:microexprnet}
Although the MM-COF representation combined with SCNN provides strong evidence of the value of dual-phase motion and magnitude modulation, its reliance on handcrafted coefficients and thresholds constrains adaptability across subjects and datasets. Furthermore, feature construction and classification remain decoupled, preventing joint optimization. To overcome these limitations, we extend MM-COF into an end-to-end trainable architecture, termed Fushion Motion Attention Network (FMANet), which embeds phase fusion and magnitude modulation directly into differentiable modules. Specifically, a Phase-Aware Consensus Fusion Block (FFB) enables adaptive pixel-wise weighting of onset and offset features, and a Soft Motion Attention Block (SMAB) provides learnable modulation of salient regions. Combined with a shallow CNN backbone, FMANet jointly optimizes feature fusion, modulation, and classification in a unified pipeline.

\subsection{Phase-Aware Consensus Fusion Block}
\begin{figure}[h]
    \centering
    \includegraphics[width=1\linewidth]{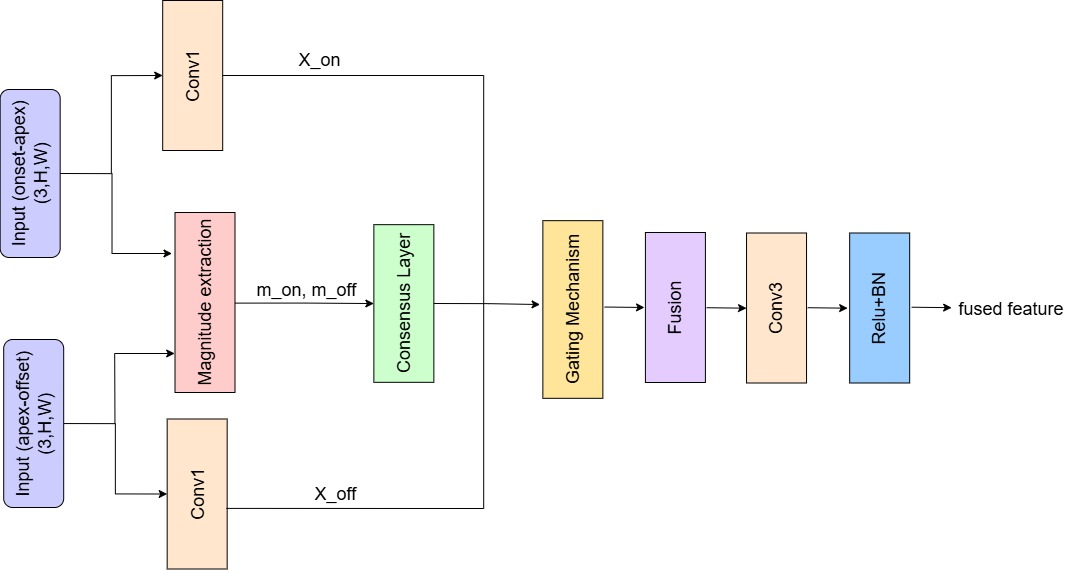}
    \caption{Consensus Fusion Block Architecture.}
    \label{fig:FFB}
\end{figure}
The Consensus Fusion Block extends the original dual-phase combination by replacing fixed weighting with an adaptive, data-driven strategy. Instead of enforcing equal or manually tuned contributions from onset-apex and apex-offset flows, FFB estimates local consensus maps based on motion strength and directional similarity. These maps guide a gating function that dynamically balances contributions from both phases at each spatial location. In this way, it adaptively combines feature representations from the onset-to-apex \((I_{on})\) and apex-to-offset\((I_{off})\) phases based on learned consensus cues.  

First, the input optical flows for each phase, \(I_{on}, I_{off} \in R^{B\times3\times H\times W} \), are passed through separate convolutional layers to extract intermediate feature maps, \(X_{on}\) and \(X_{off}\):

\[
    X_{on} = Conv_{on}(I_{on}), X_{off} = Conv_{off}(I_{off})
\]
where \(X_{on}\),\(X_{off} \in R^{B\times C_{mid}\times H\times W}\), and \(C_{mid}\) is the number of intermediate channels.

Concurrently, the module computes a consensus map, \(C_n\), which quantifies the agreement between the motion patterns of the two phases. This is achieved by evaluating both the average motion strength (\(S\)) and the similarity (\(sim\)) between their respective magnitude maps, \(M_{on}\) and \(M_{off}\). The strength \(S\) is defined as the mean magnitude:
\[
S = \frac{1}{2}(M_{on} +M_{off})
\]
The similarity is calculated as a normalized inverse of the absolute difference between magnitudes:
\[
sim = 1 - \frac{|M_{on} - M_{off}|}{M_{on} + M_{off} + \epsilon}
\]
where \(\epsilon\) is a small constant to ensure numerical stability. The consensus map \(C_n\) is then derived by combining strength and similarity, followed by max-normalization across the spatial dimensions: 
\[
C'_n = max(0,S\odot sim) 
\]
\[
C_n = \frac{C'_n}{max_{H,W}(C'_n) + \epsilon}
\]
here \(\odot\) denotes element-wise multiplication. This normalized map \(C_n\) highlights spatial regions where both phases exhibit strong and consistent motion.

To fuse the features, a gating mechanism is employed. An average feature map \(A = \frac{1}{2}(X_{on} + X_{off})\) is first computed and passed through a convolutional layer followed by a sigmoid activation to produce a preliminary gate. This gate is then modulated by the consensus map raised to the power of a hyperparameter \(\theta\), which controls the influence of the consensus score. The final gating map \(g\) is thus:
\[
g = \sigma(Conv_{graw}(A))\odot(C^\theta_{n})
\]
where \(\sigma(\cdot)\) is the sigmoid function. The final fused feature map, \(F_{raw}\) is obtained by applying this gate to linearly interpolate between the onset and offset features:
\[F_{raw} = g \odot X_{on} + (1-g)\odot X_{off}\]
This allows the network to dynamically adjust the contribution of each phase at every pixel, learning to prioritize the more discriminative motion information. The raw fused feature is then passed through a final convolutional layer, batch normalization, and a ReLU activation to produce the block's output, \(F_{fused}\).

\subsection{Soft Motion Attention}
\begin{figure}[h]
    \centering
    \includegraphics[width=1\linewidth]{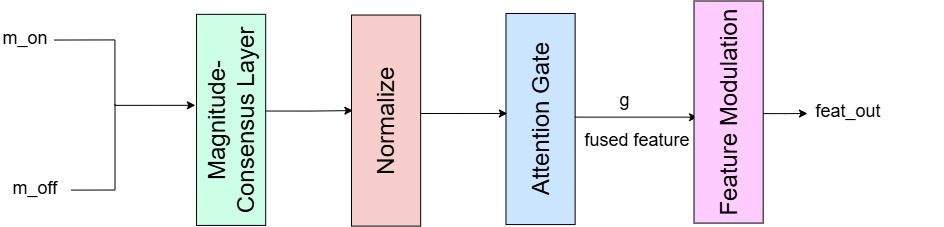}
    \caption{Soft Motion Attention Block.}
    \label{fig:smab}
\end{figure}
The second module, Soft Motion Attention Block, reformulates magnitude modulation into a learnable, soft attention mechanism. It refines the fused feature map from the FFB by selectively amplifying salient motion regions and suppressing noise. Instead of applying thresholds \((\alpha, \beta)\) and weights \((w_1, w_2)\), The attention mechanism in SMAB is guided by two motion-derived cues: the joint magnitude (capturing overall motion strength - \(M_c\)) and the directional coherence (measuring consistency between phases - \(C\)). The combined magnitude captures the joint motion intensity of both phases using a geometric mean: 
\[M_c = \sqrt{M_{on}\odot M_{off} + \epsilon}\]
The coherence map measures the temporal consistency between the phases, defined by an exponential function of their magnitude difference, with a fixed temperature parameter \(\tau\):
\[
C = exp(-\frac{|M_{on}- M_{off}|}{\tau})
\]
To ensure that these cues are scale-invariant, they are normalized across their spatial dimensions using Layer Normalization, yielding \(M_{norm} = LNorm(M_c)\) and \(C_{norm}=LNorm(C)\). These normalized maps are then concatenated and processed by a lightweight sequence of depth-wise and point-wise convolutions to generate a spatial attention gate, \(g_{attn}\).
\[
F_{in} = cat([M_{norm}, C_{norm}])
\]
\[
g_{attn} = \sigma(Conv_{pw}(ReLU(Conv_{dw}(F_{in}))))
\]
This learnable gating function allows the model to infer the importance of each spatial location from the combined motion dynamics.

Finally, the attention gate is applied to the input feature map \(F_{fused}\) from the FFB using a residual connection. This enhances the features by re-weighting them according to motion saliency: 
\[
F_{out} = F_{fused} \odot(0.5+g_{attn})
\]
The addition of 0.5 centers the modulation factor around 1.0, creating a stable residual-like connection that allows the attention to either amplify or diminish features from their original state. The output \(F_{out}\) is a motion-aware feature map, ready for the final classification backbone.

% After normalization, these descriptors are processed by lightweight depthwise and pointwise convolutions to produce an attention mask. This mask selectively emphasizes salient regions of the fused feature map while down-weighting noise or irrelevant motion, effectively generalizing the modulation step of MM-COF in a fully differentiable manner.

% The first module, FFB, extends the weighted summation used in MM-COF. Instead of assigning fixed coefficients, FFB computes feature maps from both onset and offset flows and generates an adaptive gating map. This gating is guided by two cues: (i) the strength of motion in each phase and (ii) the similarity between them. The gating value is then applied to blend onset and offset features at each spatial location. In this way, FFB automatically learns whether the onset or offset carries more discriminative information, providing a dynamic alternative to MM-COF’s static weighting.

\subsection{SCNN Backbone}
The final component of FMANet is the classification backbone. Here, we directly adopt the SCNN introduced in section~\ref{s: classification} as a lightweight yet effective feature extractor. Its shallow depth and small convolutional kernels are particularly suitable for capturing localized facial motion in MER, where training data are limited. Unlike the standalone SCNN used with MM-COF inputs, however, the backbone in FMANet is fully integrated into the pipeline. It receives features that have already been adaptively fused and modulated, ensuring that the learned representation is optimized end-to-end for recognition.

Bringing these modules together, FMANet mirrors the design logic of MM-COF and SCNN but removes handcrafted elements and replaces them with learnable mechanisms. Phase fusion FFB generalizes dual-phase weighting, motion modulation SMAB replaces threshold-based magnitude filtering, and the SCNN backbone provides efficient classification. This integration offers three key advantages: (i) adaptivity to subject-specific motion patterns, (ii) robustness against noise and irrelevant movements, and (iii) improved generalization through end-to-end optimization.

In summary, FMANet inherits the interpretability of MM-COF and the efficiency of SCNN, while overcoming their limitations by embedding both into a single, data-driven framework for micro-expression recognition.

% \section{Experiment and Results}
% \label{s: experiment and results}
% %Trình bày sơ qua về kịch bản thử nghiệm
% To evaluate the effectiveness and performance of the proposed method for Micro-Expression Recognition (MER), a series of experiments were conducted with the objective of assessing its accuracy and efficiency in recognizing micro-expressions. The experiments were designed with four primary objectives. 

% \begin{itemize}
%     \item First, an improved optical flow representation tailored for MER, referred to as MM-COF, was developed and evaluated through a visual comparison with existing methods. This evaluation highlights the ability of MM-COF to enhance the quality of optical flow representations, thereby improving MER performance. 
%     \item Second, the influence of threshold parameters on the accuracy of the proposed MER model was systematically analyzed. By varying these parameters, optimal values were identified, and their impact on recognition accuracy was assessed. 
%     \item Third, a comparative analysis of MER performance was performed using three different optical flow techniques: Single Optical Flow, Combined Optical Flow, and the proposed MM-COF. This analysis provides insights into the relative effectiveness of each technique in micro-expression recognition. 
%     \item Finally, the proposed MM-COF method was compared against state-of-the-art optical flow-based methods to further validate its robustness and superiority in enhancing MER performance.
% \end{itemize}

\section{Experiment and Results}
\label{s: experiment and results}
% To evaluate the effectiveness of our proposed MM-COF and SCNN for MER, we conducted a series of experiments focusing on their individual contributions and combined performance. The experiments were designed with three primary objectives:

% \begin{itemize}
% \item First, we assessed the effectiveness of using MM-COF, an improved optical flow representation tailored for MER. Through visual comparisons and quantitative evaluations, we demonstrated its superiority over existing optical flow methods in capturing subtle micro-expression movements.
% \item Second, we evaluated our proposed shallow CNN architecture, designed to efficiently process optical flow features while addressing the data limitations in MER. By comparing SCNN with conventional DCNN, we highlighted its effectiveness in extracting meaningful representations.
% \item Finally, we evaluated the effectiveness of the overall MER approach proposed in this paper, specifically the integration of MM-COF and Shallow CNN into a complete MER pipeline. The combination of MM-COF’s enhanced motion representation with SCNN’s optimized feature extraction led to a significant improvement in MER performance, highlighting the complementary nature of these two components.
% \end{itemize}

\subsection{Dataset}
The datasets used in this study are CASME-II~\cite{casme}, SAMM~\cite{samm} and MMEW~\cite{ben2021video}, both widely recognized as standard benchmarks in MER research due to their high-quality samples. 

% \begin{itemize}
%     \item \textbf{CASME-II}: Developed by the Chinese Academy of Sciences, the dataset consists of a total of 255 videos collected from 26 participants. It categorizes videos into seven emotion groups: others (99 videos), disgust (63 videos), happiness (32 videos), repression (27 videos), surprise (25 videos), sadness (7 videos), and fear (2 videos).
%     \item \textbf{SAMM}: It includes 159 spontaneous expression videos collected from 32 participants. The dataset categorizes expressions into 8 types as follows: anger (57 videos), happiness (26 videos), other (26 videos), surprise (15 videos), contempt (12 videos), disgust (9 videos), fear (8 videos) and sadness (6 videos).
%     \item \textbf{MMEW}: The MMEW dataset includes 300 spontaneous micro-expression samples collected in more naturalistic settings. Videos were recorded at 90 fps and annotated with senven emotion categories: happiness (36 videos), anger (8 videos), surprise (89 videos), disgust (72 videos), fear (16 videos), sadness (13 videos), others (66 videos). MMEW introduces more variation in head pose and lighting, making it a challenging benchmark for generalization.
% \end{itemize}

On the CASME-II and SAMM datasets, following common practice, categories with fewer than 10 samples are excluded, resulting in a 5-class evaluation protocol. In addition, a 3-class protocol is constructed by merging expressions into \textit{positive} (happiness), \textit{negative} (disgust, repression, sadness, fear), and \textit{surprise}, while discarding the “others” category. Meanwhile, for the MMEW dataset, we adopt multiple label configurations (3-, 5-, 6-, and 7-class) to examine the robustness of the proposed method under different levels of class granularity and imbalance, with detailed results presented in Section~\ref{sec:results}.
% For a balanced evaluation, emotion labels with fewer than 10 instances were excluded. Consequently, the CASME-II dataset was reduced to five labels: surprise, happiness, disgust, other, and repression, while the SAMM dataset was refined to surprise, happiness, contempt, other, and anger.

\textbf{Data augmentation}

To enhance the performance and generalization ability of the model, we applied data augmentation techniques to the micro-expression dataset. The specific transformations applied include:
\begin{itemize}
    \item \textit{Horizontal Flip}: Generates variations of the images by horizontally flipping them, helping the model avoid dependency on specific face orientations.
    \item \textit{Rotation}: Images are rotated at different angles, including small rotations (±5°) and larger rotations (±10°), to increase the model's ability to learn expressions without being influenced by the camera's perspective.
    % \item \textit{Normalization and Resizing}: All images are resized to 224x224 pixels and normalized to match modern deep learning models.
\end{itemize}
% These transformations are systematically applied during the training data preparation. Specifically, each original image is augmented into 10 distinct variations using predefined transformations. This not only increases the dataset size but also ensures that each image is utilized in diverse contexts, enhancing the diversity of the training dataset. As a result, the training dataset size after augmentation becomes 10 times larger than the original dataset, providing the model with diverse learning scenarios.

\subsection{Evaluation Protocol}
The evaluation follows the Leave-One-Subject-Out (LOSO) cross-validation protocol, which is commonly used in the MER field to assess subject-independent generalization. In each iteration, the samples from one subject are reserved for testing, while the remaining data are used for training.

\subsection{Evaluation Metrics}

These metrics are employed to provide a comprehensive assessment of model performance:

\begin{itemize}
    \item Accuracy: Overall ratio of correctly classified samples to the total number of samples:  
    \begin{equation}
    \label{eq:acc}
        \textit{Accuracy} = \frac{\text{Correct Predictions}}{\text{Total Samples}}.
    \end{equation}

    \item Unweighted F1 (UF1): Average of per-class F1-scores, ensuring equal contribution from each class:  
    \begin{equation}
    \label{eq:uf1}
        \textit{UF1} = \frac{1}{C} \sum_{i=1}^{C} 
        \frac{2 \cdot \text{Precision}_i \cdot \text{Recall}_i}{\text{Precision}_i + \text{Recall}_i},
    \end{equation}
    where \(C\) is the number of classes.

    \item Unweighted Average Recall (UAR): Mean of recall across all classes:  
    \begin{equation}
    \label{eq:uar}
        \textit{UAR} = \frac{1}{C} \sum_{i=1}^{C} 
        \frac{\text{TP}_i}{\text{TP}_i + \text{FN}_i},
    \end{equation}
    where \(\text{TP}_i\) and \(\text{FN}_i\) denote true positives and false negatives of class \(i\).  
\end{itemize}

\begin{figure}[h]
    \centering
    \includegraphics[width=0.9\linewidth]{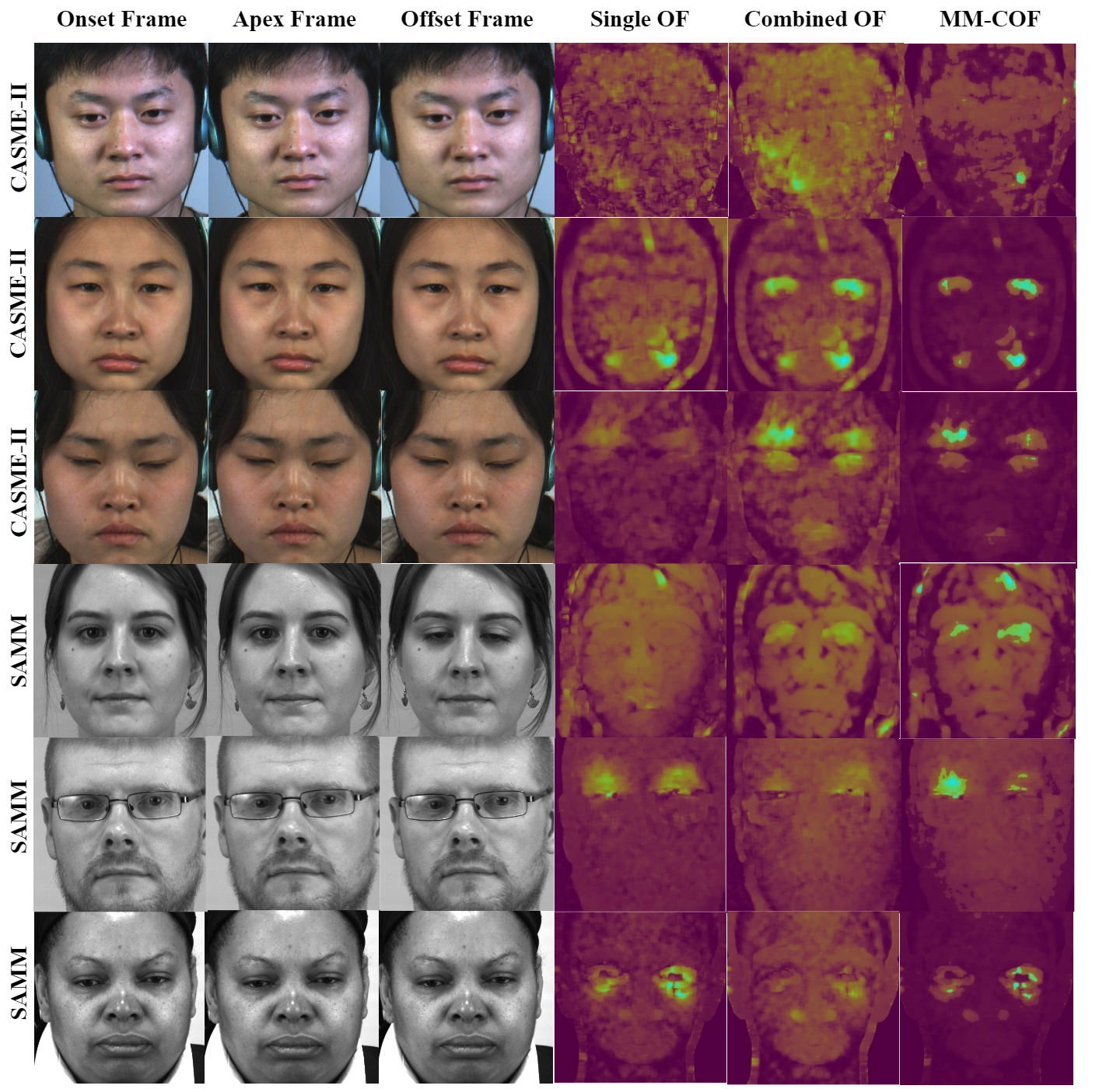}
    \caption{Onset frame, Apex frame, Offset frame, Single Optical Flow, Combined Optical Flow and Magnitude-Modulated Combined Optical Flow (ordered from left to right) on CASME-II and SAMM datasets. \textit{(Best viewed in color)}}
    \label{fig:visual01}
\end{figure}

\subsection{Results}
\label{sec:results}
\subsubsection{Visual representation results}
Fig. \ref{fig:visual01} demonstrate the differences between Single Optical Flow, Combined Optical Flow, and our proposed Magnitude-Modulated Combined Optical Flow, along with the onset, apex, and offset frames. By applying the threshold, our method focuses on the regions with significant motion, filtering out irrelevant or less dynamic areas. Visual inspection shows that the MM-COF captures more detailed transitions in facial expressions, particularly those that are crucial for recognizing micro-expressions. This selective focus on key dynamic regions likely contributes to the observed performance improvement in Micro-Expression Recognition compared to Single Optical Flow and Combined Optical Flow.

\begin{figure}[h]
    \centering
    \includegraphics[width=0.9\linewidth]{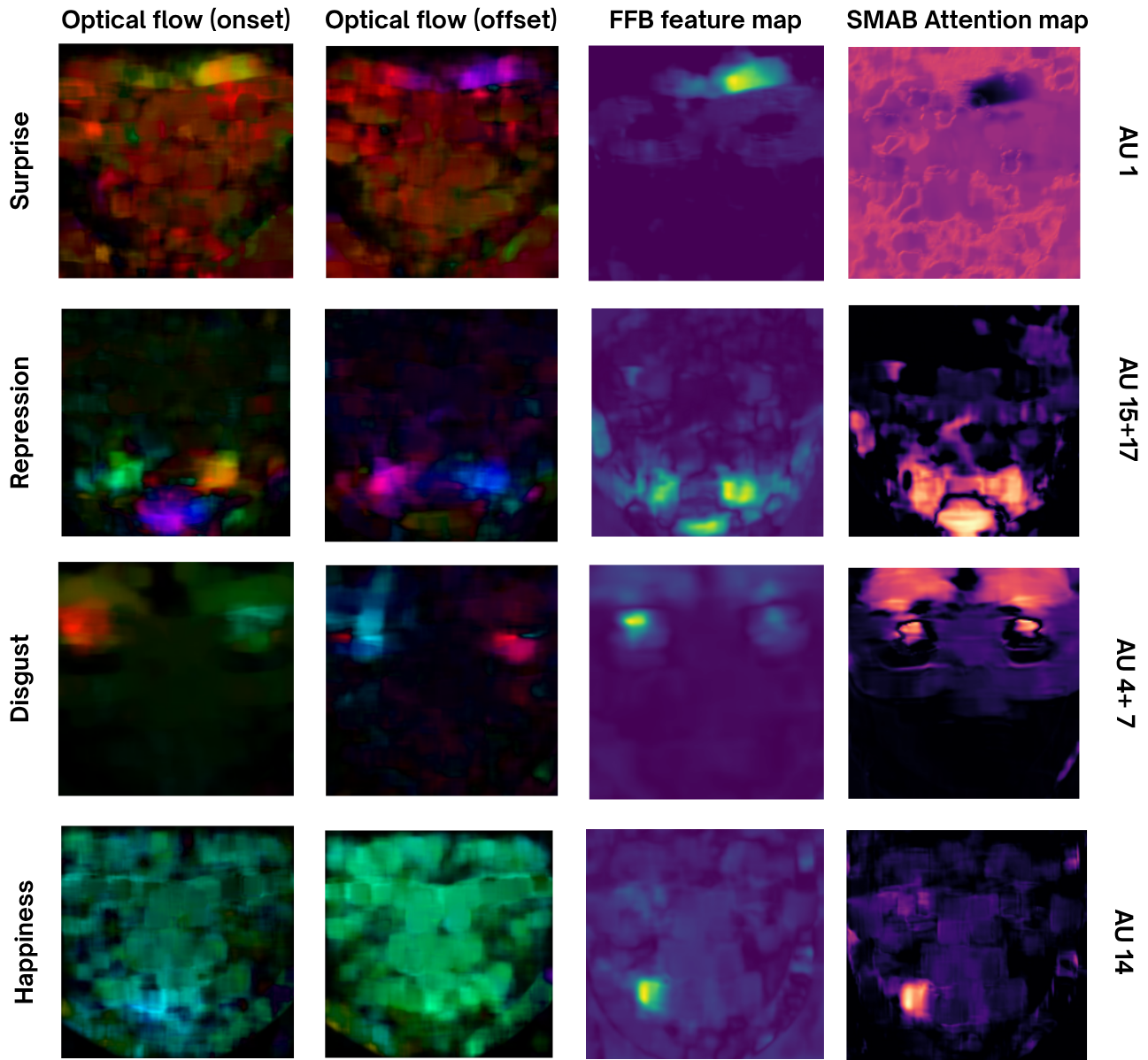}
    \caption{Optical flow and the corresponding FFB Features map and SMAB Attention map. \textit{(Best viewed in color)}}
    \label{fig:visual02}
\end{figure}

Figure \ref{fig:visual02} visualizes the outputs of the components within the proposed FMANet architecture. From left to right, the images represent: the optical flow during the onset phase, the optical flow during the offset phase, the corresponding feature map generated by the FFB (Phase-Aware Consensus Fusion Block), and the attention map obtained from the SMAB (Soft Motion Attention Block).

The results indicate that both the FFB and SMAB blocks successfully fulfill their intended purpose. The FFB effectively combines homogeneous regions from the two motion phases, and the SMAB concentrates on the critical regions within the FFB's feature representation. Specifically, compared to the original optical flow representations, the motion features generated by the FFB are more effective at reducing noise and focusing on significant areas. For example, in the "Surprise" and "Happiness" expressions, the movements are subtle and tend to produce considerable noise in the optical flow representation. However, our method successfully distills the core motion information by unifying the two motion phases, and the SMAB effectively targets these key regions. Furthermore, the results demonstrate that with the representations generated by the FFB and SMAB, the corresponding Action Units can be observed and identified much more easily and accurately than with the original optical flow representation.
\begin{table*}[ht]
\centering
\caption{Comparison of micro-expression recognition performance on the CASME II and SAMM datatsets for the state-of-the-art method (5 classes)}
\begin{tabular}{clccccccc}
\toprule
\multirow{3}{*}{No} & \multirow{3}{*}{Methods} & \multirow{3}{*}{Year} &
\multicolumn{3}{c}{\textbf{SAMM}} &
\multicolumn{3}{c}{\textbf{CASME-II}} \\
\cmidrule(lr){4-6}\cmidrule(lr){7-9}
 & &  & Acc & UF1 & UAR & Acc & UF1 & UAR \\
\midrule
1 & FHOFO \cite{7971947} 
& 2019 
& - & - & - & 55.86 & 0.5197 & - \\
2 & AU-GACN \cite{10.1145/3394171.3414012} 
& 2020
& 52.3 & 0.357 & - & 56.1 & 0.394 & - \\
3 & Micro-Attention \cite{WANG2020354}
& 2020
& 48.5 & 0.402 & - & 65.9 & 0.539 & - \\
4 & LGCcon \cite{9250649} 
& 2021
& 40.9 & 0.34 & - & 65.2 & 0.64 & - \\
5 & KFC-MER \cite{9428407} 
& 2021
& 63.24 & 0.57 & - & 72.76 & 0.738 & - \\
6 & Dynamic \cite{9064925} 
& 2022
& - & - & - & 72.61 & 0.67 & - \\
7 & MER-Supcon \cite{ZHI202225} 
& 2022
& 67.65 & 0.625 & - & 73.58 & 0.728 & - \\
8 & KPCANet \cite{10.1145/3607829.3616444} 
& 2023
& 63.83 & 0.522 & - & 70.46 & 0.659 & - \\
9 & AU GCN \cite{augcn24} 
& 2024
& 79.82 & 0.757 & - & 81.85 & 0.776 & - \\
10 & MESTI-MEGANet \cite{nguyen2025mestimeganetmicroexpressionspatiotemporalimage} 
& 2025
& 80.88 & 0.791 & 0.803 & 82.04 & 0.779 & 0.786 \\
11 & MELLM \cite{zhang2025mellmexploringllmpoweredmicroexpression} 
& 2025
& - & - & - & 64.34 & 0.485 & 0.534 \\
12 & SODA4MER \cite{soda}
& 2025
& 80.30 & 0.789 & - & 84.18 & 0.814 & - \\
\midrule
13 & \textbf{MM-COF+SCNN (ours)}
& 2025
& 66.67 &  0.588 & 0.595 & 62.30 & 0.597 & 0.621 \\
14 & \textbf{FMANet (ours)} 
& 2025
& 84.56	&0.810&	0.799 & 73.71 & 0.659 & 0.677 \\
\bottomrule
\end{tabular}
\label{tab:sota5class}
\end{table*}

\begin{table*}[ht]
\centering
\caption{Comparison of micro-expression recognition performance on the CASME II and SAMM datasets for the state-of-the-art method (3 classes)}
\begin{tabular}{clccccccc}
\toprule
\multirow{3}{*}{No} & \multirow{3}{*}{Methods} & \multirow{3}{*}{Year} &
\multicolumn{3}{c}{\textbf{SAMM}} &
\multicolumn{3}{c}{\textbf{CASME-II}} \\
\cmidrule(lr){4-6}\cmidrule(lr){7-9}
 & &  & Acc & UF1 & UAR & Acc & UF1 & UAR \\
\midrule
1 & STSTNet \cite{stst} 
& 2019
& - & 0.658 & 0.681 & - & 0.838 & 0.868 \\
2 & AU-GACN \cite{10.1145/3394171.3414012} 
& 2020
& 70.2 & 0.433 & - & 71.2 & 0.355 & - \\
3 & MiMaNet \cite{mima} 
& 2021
& 76.7 & 0.764 & - & 79.9 & 0.759 & - \\
4 & GEME \cite{NIE202113} 
& 2021
& - & 0.584 & 0.545 & - & 0.883 & 0.879 \\
5 & MER-Supcon \cite{ZHI202225} 
& 2022
& 81.20 & 0.713 & - & 89.65 & 0.881 & - \\
6 & FRL-DGT  \cite{frldgt}
& 2023
& - & 0.772 & 0.758 & - & 0.919 & 0.903 \\
7 & MOL \cite{mol}
& 2025
& 88.36 & 0.827 & - & 91.26 & 0.889 & - \\
8 & MPFNet \cite{ma2025mpfnetmultipriorfusionnetwork}     
& 2025 
& 85.0 & 0.856 & -- & 89.70 & 0.898 & -- \\
\midrule
9 & \textbf{MM-COF+SCNN (ours)}
& 2025
& 78.9 & 0.735 & 0.755 & 82.07 & 0.796 & 0.885 \\
10 & \textbf{FMANet (ours)} 
& 2025
&  88.24	&0.841&	0.809& 87.30 & 0.807 & 0.805 \\
\bottomrule
\end{tabular}
\label{tab:sota3class}
\end{table*}
\subsubsection{Comparison with State-of-the-arts}

Table~\ref{tab:sota5class} and Table~\ref{tab:sota3class} report results under the 5-class and 3-class settings on CASME-II and SAMM. On SAMM, FMANet consistently delivers state-of-the-art performance. For example, while conventional CNN-based models such as KFC-MER~\cite{9428407} or MER-SupCon~\cite{ZHI202225} remain below 70\% accuracy, and even recent graph-based approaches like AUGCN~\cite{10.1145/3607829.3616444} achieve around 80\%, FMANet pushes the performance to 84.56\% accuracy with superior UF1 and UAR. This improvement highlights the importance of explicitly capturing dual-phase motion dynamics, which conventional frame-level or single-phase representations fail to model effectively. In the 3-class protocol, a similar trend can be observed: FMANet maintains both high accuracy and balanced metrics, whereas methods such as MPNet~\cite{ma2025mpfnetmultipriorfusionnetwork} or MOL achieve competitive accuracy but show less consistency across UF1 and UAR. This indicates that FMANet not only achieves strong recognition rates but also generalizes better under imbalanced distributions, which are common in SAMM.

On CASME-II, the situation is more challenging due to shorter sequences and lower-intensity expressions. Transformer- and GCN-based methods (e.g., AUGCN~\cite{10.1145/3394171.3414012}, MESTI-MEGANet~\cite{nguyen2025mestimeganetmicroexpressionspatiotemporalimage}) report slightly higher accuracies, but FMANet achieves comparable performance with more stable UF1/UAR. This suggests that our phase-aware representations remain effective even when the motion signal is weak, though further gains may require specialized mechanisms (e.g., temporal magnification or stronger attention).

Overall, FMANet advances the state-of-the-art on SAMM and remains highly competitive on CASME-II. Compared with prior CNN- and GCN-based methods, our framework demonstrates that explicit modeling of onset–apex–offset dynamics provides tangible benefits for micro-expression recognition, especially in datasets with richer temporal information.

% \begin{table*}[t]
%     \centering
%     \caption{Comparison of micro-expression recognition performance on the MMEW dataset}
%     \setlength{\tabcolsep}{6pt}
%     \renewcommand{\arraystretch}{1.15}
%     \begin{tabular}{clccccccc}
%         \toprule
%         \multirow{3}{*}{\textbf{No}} & \multirow{3}{*}{\textbf{Methods}} 
%         & \textbf{Year}
%         & \multicolumn{6}{c}{\textbf{MMEW}} \\
%         \cmidrule(lr){4-9}
%         & && \multicolumn{3}{c}{\textbf{4 classes}} 
%           & \multicolumn{3}{c}{\textbf{7 classes}} \\
%         \cmidrule(lr){4-6}\cmidrule(lr){7-9}
%         & & & \textbf{Acc (\%)} & \textbf{UF1 (\%)} & \textbf{UAR (\%)} 
%          &  \textbf{Acc (\%)} & \textbf{UF1 (\%)} & \textbf{UAR (\%)} \\
%         \midrule
       
%         9 & \textbf{MM-COF + SCNN (Ours)}  & 2025                                  & \textbf{--} & \textbf{--} & \textbf{--} & \textbf{--} & \textbf{--} & \textbf{--} \\
%         9 & \textbf{FMANEt (Ours)}     & 2025                               & \textbf{85.17} & \textbf{83.34} & \textbf{83.66} & \textbf{75.67} & \textbf{65.09} & \textbf{64.51} \\
%         \bottomrule
%     \end{tabular}
%     \label{tab:mmew_sota_4vs7}
% \end{table*}

To further assess the generalization ability of FMANet beyond these controlled datasets, we extend our experiments to the more diverse MMEW corpus. The MMEW dataset originally contains seven emotion categories; however, following previous works, we also report results under different label configurations (3, 5, 6, and 7 classes) (see Table~\ref{tab:mmew_sota_classes}). This allows us to evaluate the robustness of the proposed FMANet when dealing with different levels of class granularity and label imbalance. FMANet consistently surpasses CNN-based baselines (e.g., ShuffleNet~\cite{liu2022micro}, LD-FMEN~\cite{NI2023110729}) and graph-based approaches (e.g., TDSGCN, GCN), particularly in the 3- and 5-class settings where it achieves more balanced UF1 and UAR. When extended to 6- and 7-class protocols, FMANet still outperforms recent transformer or GCN-based models, though performance drops slightly due to the challenge of fine-grained categories and class imbalance. These results confirm that FMANet is robust across varying levels of granularity, with strong potential for practical deployment in reduced-class scenarios while maintaining competitiveness on the full 7-class evaluation.

\begin{table}[h]
    \centering
    \caption{Comparison of micro-expression recognition performance on the MMEW dataset across different class settings. Results are reported in Acc, UF1, and UAR (\%).}
    \begin{tabular}{clcccccc}
        \toprule
        \textbf{No}& \textbf{Methods}
        & \textbf{Year}
        & \textbf{Acc} & \textbf{UF1} & \textbf{UAR} \\
        \midrule

        % --------- 3 classes ------------
        \multicolumn{6}{c}{\textbf{3 classes}} \\
        \midrule
        1 & ShuffleNet~\cite{liu2022micro} &2022 & 69.81	& 73.18 & - \\
        2 &LD-FMEN~\cite{NI2023110729} & 2023 & 88.23 &	87.87	&87.76 \\
        3 & EDMDBN~\cite{MA2025166} & 2025 & 92.70 &	92.16 & - \\
        4 & \textbf{FMANet (Ours)} & 2025 & \textbf{85.47} & \textbf{82.3} & \textbf{82.73} \\

        % % --------- 4 classes ------------
        % \midrule
        % \multicolumn{6}{c}{\textbf{4 classes}} \\
         % \midrule
        % 5 & FMANet (Ours)        & 2025 & \textbf{85.17} & \textbf{83.34} & \textbf{83.66} \\

        % --------- 5 classes ------------
        \midrule
        \multicolumn{6}{c}{\textbf{5 classes}} \\
        \midrule
         5& TDSGCN~\cite{tang2022transferring} & 2022 & 72.7& -&-  \\
         6& GCN~\cite{TANG2024106421} & 2024 & 70.3 & -&-  \\
          7& \textbf{FMANet (Ours)} & 2025 & \textbf{81.90} & \textbf{72.31} & \textbf{71.74} \\

        % --------- 6 classes ------------
        \midrule
        \multicolumn{6}{c}{\textbf{6 classes}} \\
        \midrule
        8& CoDER~\cite{li2024counterfactual} & 2024 & 74.8 &- &- \\
        9& MERba-DGCM~\cite{mao2025merba} & 2025 &75.2 &- & -\\
        10 & \textbf{FMANet (Ours)} & 2025 & \textbf{83.33} & \textbf{70.15} & \textbf{64.71} \\

        % --------- 7 classes ------------
        \midrule
        \multicolumn{6}{c}{\textbf{7 classes}} \\
        \midrule
        11& Sparse Transformer~\cite{zhu2022sparse} & 2022  &73.93& - &- \\
        12 & \textbf{FMANet (Ours)} & 2025 & \textbf{75.67} & \textbf{65.09} & \textbf{64.51} \\

        \bottomrule
    \end{tabular}
    \label{tab:mmew_sota_classes}
\end{table}

\subsection{Ablation studies}
\label{s: ablation studies}

\subsubsection{Evaluation of input modalities}

% We evaluate the effectiveness of different input modalities under two settings: (i) single-stream baselines with VGG19 and the proposed SCNN, and (ii) the dual-stream FMANet that jointly models onset-apex and apex-offset phases. 

Table~\ref{tab:dual_stream} investigates the impact of different input modalities in FMANet. A consistent trend emerges: the phase-aware optical flow stream alone achieves highly competitive results across both CASME-II and SAMM, confirming that even without additional modulation it already captures the critical dynamics of micro-expressions. The proposed magnitude-modulated optical flow (MM-OF) provides complementary cues and can improve class balance in certain cases, yet its overall advantage over plain optical flow is not always pronounced. This observation suggests two key insights: first, that optical flow itself is a strong and reliable modality when modeled in a dual-phase manner; and second, that MM-OF, while promising, requires further refinement to consistently surpass its baseline counterpart. Taken together, these findings highlight optical flow as the core driver of FMANet’s success, with MM-OF serving as a meaningful extension that broadens the potential for handling imbalance and subtle variations.

\begin{table}[h]
    \centering
    \caption{Evaluation of different input modalities in FMANet on the CASME-II and SAMM datasets. Results are reported in Acc, UF1, and UAR (\%).}
    \begin{tabular}{lcccccc}
        \toprule
        \multirow{2}{*}{Input modality} & 
        \multicolumn{3}{c}{CASME-II} & \multicolumn{3}{c}{SAMM} \\
        \cmidrule(lr){2-4} \cmidrule(lr){5-7}
        & Acc  & UF1 & UAR 
        & Acc & UF1  & UAR\\
        \midrule
        \multicolumn{7}{c}{\textbf{5 classes}} \\
        \midrule
        Optical flow   & \textbf{73.71} & 65.85 & 67.73 & \textbf{84.56} & \textbf{81.03} & \textbf{79.88} \\
        MM-OF          & 73.55 & \textbf{66.43} & \textbf{69.28} & 77.11 & 76.98 & 77.87 \\
        \midrule
        \multicolumn{7}{c}{\textbf{3 classes}} \\
        \midrule
        Optical flow   & \textbf{87.30} & \textbf{80.70} & \textbf{80.50} & \textbf{88.24} & \textbf{84.12} & \textbf{80.09} \\
        MM-OF          & 83.10 & 74.48 & 76.08 & 76.65 & 69.21 & 70.52 \\
        \bottomrule
    \end{tabular}
    \label{tab:dual_stream}
\end{table}

% \begin{table}[ht]
%     \centering
%     \caption{Evaluation of different input modalities under single-stream (SCNN and VGG19) and dual-stream (FMANet) settings on the CASME-II and SAMM datasets.}
%     \begin{tabular}{llcccccc}
%         \toprule
%         \multirow{2}{*}{Model} & \multirow{2}{*}{Input} & 
%         \multicolumn{3}{c}{CASME-II} & \multicolumn{3}{c}{SAMM} \\
%         \cmidrule(lr){3-5} \cmidrule(lr){6-8}
%         & & Acc & UF1 & UAR & Acc & UF1 & UAR \\
%         \midrule
%         \multirow{2}{*}{\textbf{FMANet (2 input)}} 
%         & Optical flow   & \textbf{73.71} & 65.85 & 67.73 & - & - & - \\
%         & MM-OF  & 73.55 & \textbf{66.43} & \textbf{69.28} & 77.11 & 76.98 & 77.87 \\
%         \midrule
%         \multirow{3}{*}{\textbf{SCNN (1 input)}} 
%         & Apex frame  & - & - & - & - & - & - \\
%         & Optical flow  & - & - & - & - & - & - \\
%         & MM-OF  & - & - & - & - & - & - \\
%         & MM-COF            & 72.65 & - & - & 84.56 & - & - \\
%         \bottomrule
%         \multirow{3}{*}{VGG19 (1 input)} 
%         & Apex frame  & - & - & - & - & - & - \\
%         & Optical flow  & - & - & - & - & - & - \\
%         & MM-OF  & - & - & - & - & - & - \\
%         & MM-COF            & 72.65 & - & - & 84.56 & - & - \\
%         \bottomrule
%     \end{tabular}
%     \label{tab:input_eval}
% \end{table}

\subsubsection{Module contribution analysis in FMANet}

Table~\ref{tab:ablation_samm} reports the effect of individual components on recognition performance on SAMM dataset. Using SCNN alone provides the baseline. Adding either the Feature Fusion Block (FFB) or the Soft Motion Attention Block (SMAB) consistently improves UF1 and UAR, demonstrating their ability to enhance discriminative motion features and better handle class imbalance. When only FFB is added to SCNN, the model achieves a notable gain on the SAMM dataset, highlighting the importance of multi-phase motion fusion. Similarly, incorporating SMAB improves performance by selectively emphasizing subtle motion cues. The full model, combining SCNN with both FFB and SMAB, achieves the best overall results across CASME-II and SAMM. These findings confirm that each component plays a complementary role: FFB enriches motion integration, SMAB improves motion saliency modeling, and SCNN provides a strong backbone. Together, they yield the most robust and balanced configuration for MER.

\begin{table}[h]
    \centering
    \caption{Ablation study of FMANet components on the SAMM dataset (5 classes)}
    \setlength{\tabcolsep}{4pt}
    \begin{tabular}{cccccc}
        \toprule
        \multicolumn{3}{c}{Components} & \multicolumn{3}{c}{\textbf{SAMM}} \\
        \cmidrule(lr){1-3} \cmidrule(lr){4-6}
        FFB & SMAB & SCNN & Acc & UF1 & UAR \\
        \midrule
       \checkmark & \checkmark & \checkmark & \textbf{84.56} & \textbf{81.03} & \textbf{79.88} \\
                  & \checkmark & \checkmark & 72.06 & 62.39 & 61.55 \\
       \checkmark &            & \checkmark & 72.06 & 63.67 & 63.57\\
                  &            & \checkmark & 69.12 & 60.04 &  59.37 \\
        \bottomrule
    \end{tabular}
    \label{tab:ablation_samm}
\end{table}
\begin{table*}[h]
\centering
\begin{minipage}{0.48\textwidth}
\caption{MER Accuracy Using Thresholding for MM-COF on the CASME-II Dataset}
\begin{tabular}{cccc}
\hline
\textbf{$\alpha$} & \textbf{$\beta$} & \text{VGG19 (\%)} & \text{SCNN (ours) (\%)} \\
\hline
\multicolumn{4}{c}{\textbf{Manual Thresholding}}  \\
\hline
% $0.2$ & $1$ & 51.63 & - \\
$0.2$ & $0.8$ & 56.56 & 68.03 \\
$0.2$ & $0.9$ & 56.97 & 67.21 \\
$0.2$ & $1.0$ & 52.87 & 65.98 \\
$0.3$ & $1.0$ & 54.10 & 65.98 \\
$0.3$ & $1.3$ & 55.74 & 66.39 \\
$0.3$ & $1.6$ & 55.74 & 65.57 \\
$0.3$ & $1.8$ & 54.47 & 65.98 \\
$0.4$ & $1.2$ & 57.81 & 65.16 \\
$0.4$ & $1.3$ & 56.16 & 68.03 \\
$0.4$ & $1.4$ & 59.01 & 66.39 \\
$0.4$ & $1.5$ & 58.64 & 64.34 \\
$0.4$ & $1.6$ & 58.61 & 65.16 \\
$0.4$ & $1.7$ & 59.86 & 63.11 \\
$0.4$ & $1.8$ & 58.20 & 66.39 \\
$0.4$ & $1.9$ & 58.62 & 68.03 \\
\textbf{$0.5$} & \textbf{$1.0$} & \textbf{61.48} & 67.21 \\
$0.5$ & $1.3$ & 58.23 & 67.21 \\
$0.5$ & $1.4$ & 57.81 & 63.52 \\
$0.5$ & $1.5$ & 58.64 & 64.75 \\
$0.5$ & $1.6$ & 58.22 & 63.11 \\
$0.5$ & $1.7$ & 58.62 & 66.39 \\
$0.5$ & $1.8$ & 58.21 & \textbf{68.85} \\
\hline
\multicolumn{4}{c}{\textbf{Adaptive Thresholding}}  \\
\hline
 $\alpha_{adaptive}$ & $\beta_{adaptive}$ & \textbf{63.93} & \textbf{70.08} \\
\hline
\end{tabular}
\label{tab:result1}
\end{minipage}%
\hfill
\begin{minipage}{0.48\textwidth}
\centering
\caption{MER Accuracy Using Thresholding for MM-COF on the SAMM Dataset}
\begin{tabular}{cccc}
\hline
\textbf{$\alpha$} & \textbf{$\beta$} & VGG19 (\%) & SCNN (ours)(\%) \\
\hline
\multicolumn{4}{c}{\textbf{Manual Thresholding}}  \\
\hline
% $0.2$ & $1$ & 57.63 & - \\
\textbf{$0.2$} & \textbf{$0.8$} & \textbf{60.74} & \textbf{63.70} \\
$0.2$ & $0.9$ & 59.26 & 60.00 \\
$0.2$ & $1.0$ & 60.47 & 60.00 \\
$0.3$ & $1.0$ & 60.00 & 62.96 \\
$0.3$ & $1.3$ & 57.78 & 61.48 \\
$0.3$ & $1.6$ & 52.59 & 60.00 \\
$0.3$ & $1.8$ & 55.56 & 61.48 \\
$0.4$ & $1.2$ & 56.30 & 62.22 \\
$0.4$ & $1.3$ & 51.85 & 60.74 \\
$0.4$ & $1.4$ & 48.89 & 58.52 \\
$0.4$ & $1.5$ & 51.85 & 57.78 \\
$0.4$ & $1.6$ & 52.59 & 57.78 \\
$0.4$ & $1.7$ & 50.37 & 57.78 \\
$0.4$ & $1.8$ & 54.81 & 59.26 \\
$0.4$ & $1.9$ & 52.29 & 60.00 \\
$0.5$ & $1.0$ & 58.52 & 62.22 \\
$0.5$ & $1.3$ & 51.11 & 59.26 \\
$0.5$ & $1.4$ & 49.63 & 58.52 \\
$0.5$ & $1.5$ & 53.33 & 57.78 \\
$0.5$ & $1.6$ & 53.33 & 60.00 \\
$0.5$ & $1.7$ & 53.33 & 57.78 \\
$0.5$ & $1.8$ & 53.33 & 60.00 \\
\hline
\multicolumn{4}{c}{\textbf{Adaptive Thresholding}}  \\
\hline
 $\alpha_{adaptive}$ & $\beta_{adaptive}$  & \textbf{60.74} & \textbf{63.70} \\
\hline
\end{tabular}
\label{tab:result2}
\end{minipage}
\end{table*}

\subsubsection{Analysis of Magnitude-modulated Combined Optical Flow (MM-COF)} 

In this experiment, we employ a $k$-fold cross-validation protocol to systematically evaluate different configurations of the MM-COF input representation. The goal is to identify the most effective setting that improves the discriminative power of subtle motion features. In particular, we analyze the impact of thresholding and weighting parameters, which are designed to emphasize significant motion cues while suppressing noise. These parameters play a critical role in enhancing both the robustness and the recognition accuracy of micro-expressions. The results demonstrate which configuration of MM-COF yields the best overall performance, serving as the optimal input representation for subsequent model components.

% \subsubsection{Threshold optimization in MM-COF} \label{s: threshold variability}

\textbf{Threshold selection:} Thresholding is crucial for emphasizing key dynamic regions and filtering out irrelevant or low-motion areas. We evaluate two strategies: manual and adaptive thresholding (as shown in Table~\ref{tab:result1} and Table~\ref{tab:result2}).

\textit{Manual thresholding:} In this approach, thresholds are selected through iterative tuning. Initial values are tested over a broad range and refined based on recognition accuracy. Due to differences in motion intensity and expression duration across datasets, the optimal thresholds vary considerably. For CASME-II, higher thresholds $(\alpha=0.5, \beta=1.8)$ achieved the best accuracy of 68.85\% by filtering noise from less dynamic regions while preserving salient motion. For SAMM, lower thresholds $(\alpha=0.2, \beta=0.8)$ yielded the highest accuracy of 63.70\%, better capturing its short and subtle expressions. Although effective, manual thresholding requires extensive tuning and lacks generalizability.

\textit{Adaptive thresholding:} To address these limitations, an adaptive mechanism was introduced, dynamically adjusting thresholds per sample. This allows the model to better accommodate variations in motion magnitude and expression duration without dataset-specific tuning. As shown in Table~\ref{tab:result1} and Table~\ref{tab:result2}, adaptive thresholding improved recognition on CASME-II to 70.08\% and maintained stable performance on SAMM (63.70\%). 

\textbf{Weighting strategy:} In addition to threshold selection, the weighting parameters $w_1$ and $w_2$ play a crucial role in motion magnitude modulation by balancing enhancement and suppression effects. These parameters regulate the emphasis on dominant and subtle motion patterns, directly impacting the quality of feature extraction for micro-expression recognition. Using adaptive thresholding as the baseline, we evaluated multiple configurations of $(w_1, w_2)$ (Table~\ref{tab:weight}). The setting $w_1=2, w_2=1/2$ consistently yielded the best accuracy across both CASME-II and SAMM. Larger $w_1$ values (e.g., 3 or 4) over-amplified dominant motions and masked subtle expressions, while smaller $w_2$ values weakened enhancement, degrading recognition accuracy.

\begin{table}[h]
    \caption{Evaluation of different weight parameter settings in adaptive thresholding}
    \centering
    \begin{tabular}{cccc}
        \hline
        \multirow{2}{*}{$w_1$} & \multirow{2}{*}{$w_2$} & \multicolumn{2}{c}{SCNN (Ours) (\%)} \\
        \cline{3-4}
        & & CASME & SAMM \\
        \hline
        Fixed max & Fixed min & 65.98 & 62.22 \\
        2 & 1/2 & \textbf{70.08} & \textbf{63.70} \\
        3 & 1/3 & 68.03 & 60.74 \\
        4 & 1/4 & 65.98 & 62.22 \\
        \hline
    \end{tabular}
    \label{tab:weight}
    \vspace{2mm} % Tạo khoảng cách nhỏ trước chú thích
    \caption*{Fixed min/max: Motion values are directly assigned to their minimum and maximum instead of applying weighting factors.}
\end{table}

\textbf{Phase impact factor:} The optimization of \(\theta\) parameters aimed to balance the contributions of the onset-to-apex and apex-to-offset phases in MM-COF.  Table \ref{tab:theta} reports the performance of different \(\theta_1, \theta_2\) settings on the CASMEII and SAMM datasets using the proposed SCNN model. Equal weighting \((\theta_1= \theta_2=1)\) consistently delivers the most stable results, achieving 70.08\% on CASME-II and 63.70\% on SAMM. Assigning higher weight to either phase improves performance on one dataset but degrades it on the other, confirming that equal contribution is the optimal and balanced configuration.

\begin{table}[h]
    \caption{Evaluation of different theta parameter settings of MM-COF in adaptive thresholding on CASME-II and SAMM}
    \centering
    \begin{tabular}{cccc}
        \hline
        \multirow{2}{*}{$\theta_1$} & \multirow{2}{*}{$\theta_2$} & \multicolumn{2}{c}{SCNN (Ours) (\%)} \\
        \cline{3-4}
        & & CASME-II & SAMM \\
        \hline
        1 & 1 & \textbf{70.08} & 63.70\\
        1 & 2 & 65.57 & \textbf{65.19} \\
        2 & 1 & 66.84 & 62.96 \\
        \hline
    \end{tabular}
    \label{tab:theta}
\end{table}

\textbf{Input modalities:} Table~\ref{tab:single_stream} further compares MM-COF against conventional representations using three backbone networks: VGG19, ConvNeXtV2, and the proposed SCNN. VGG19 is included as a widely adopted baseline CNN with a deep yet generic architecture, while ConvNeXtV2 represents a more recent convolutional design optimized for large-scale visual recognition. Despite their capacities, both models are not specifically tailored for capturing subtle and low-amplitude facial motions. In contrast, SCNN demonstrates superior and more consistent performance across both datasets, highlighting the advantage of a compact, task-focused architecture for micro-expression analysis. Moreover, MM-COF paired with SCNN yields the highest single-stream results, achieving 70.08\% on CASME-II and 63.70\% on SAMM, confirming the effectiveness of motion modulation in enhancing discriminative power.

% \begin{table}[h]
%     \centering
%     \caption{Evaluation of different input modalities in the single-stream setting using VGG19, MobileNet, and SCNN on the CASME-II and SAMM datasets. Results are reported in Accuracy (\%).}
%     \begin{tabular}{lcccccc}
%         \toprule
%         \multirow{2}{*}{Input modality} 
%         & \multicolumn{3}{c}{CASME-II} 
%         & \multicolumn{3}{c}{SAMM} \\
%         \cmidrule(lr){2-4} \cmidrule(lr){5-7}
%          & VGG19 & ConvNeXTV2 & \textbf{SCNN (ours)}
%          & VGG19 & ConvNeXTV2 & \textbf{SCNN (ours)} \\
%         \midrule
%         Apex frame        & 59.59 & 60.66
%  & 63.27  & 55.15 & 57.04
%  & 54.41 \\
%         Optical flow      & 54.10 & 59.84 & 63.52  & 58.15 &  56.3 & 54.41 \\
%         COF               & 56.97 & 61.89 & 65.98  & 58.52 & 57.78 & 59.26 \\
%         \textbf{MM-COF}   & \textbf{63.93} & \textbf{64.75} & \textbf{70.08}  
%                           & \textbf{60.74} & \textbf{60} & \textbf{63.70} \\
%         \bottomrule
%     \end{tabular}
%     \label{tab:single_stream}
% \end{table}

\begin{table}[h]
    \centering
    \caption{Evaluation of different input modalities using VGG19, ConvNeXtV2, and SCNN on CASME-II and SAMM datasets (Accuracy \%).}
    \begin{tabular}{l l c c}
        \toprule
        \textbf{Input Modality} & \textbf{Model} & \textbf{CASME-II} & \textbf{SAMM} \\
        \midrule
        \multirow{3}{*}{Apex frame} 
            & VGG19        & 59.59 & 55.15 \\
            & ConvNeXtV2   & 60.66 & 57.04 \\
            & \text{SCNN (ours)} & \text{63.27} & 54.41 \\
        \midrule
        \multirow{3}{*}{Optical flow} 
            & VGG19        & 54.10 & 58.15 \\
            & ConvNeXtV2   & 59.84 & 56.30 \\
            & \text{SCNN (ours)} & \text{63.52} & 54.41 \\
        \midrule
        \multirow{3}{*}{COF} 
            & VGG19        & 56.97 & 58.52 \\
            & ConvNeXtV2   & 61.89 & 57.78 \\
            & \text{SCNN (ours)} & \text{65.98} & 59.26 \\
        \midrule
        \multirow{3}{*}{\textbf{MM-COF}} 
            & VGG19        & 63.93 & 60.74 \\
            & ConvNeXtV2   & 64.75 & 60.00 \\
            & \textbf{SCNN (ours)}  & \textbf{70.08} & \textbf{63.70} \\
        \bottomrule
    \end{tabular}
    \label{tab:single_stream}
\end{table}

\section{Conclusion}
\label{s:conclusion}

This paper has proposed the magnitude-modulated combined optical flow (MM-COF) as a principled representation for micro-expression recognition. Unlike prior optical flow approaches that mainly emphasize the onset-to-apex phase, MM-COF integrates both onset–apex and apex–offset dynamics, offering a more comprehensive description of facial motion. Combined with magnitude modulation, which adaptively enhances subtle cues while suppressing noise, MM-COF provides a robust foundation that consistently outperforms conventional flow-based inputs. Extensive ablation further confirmed that equal phase weighting and balanced magnitude modulation yield the most stable and discriminative configuration.

Building upon this representation, we introduced FMANet, an end-to-end dual-stream network that extends the MM-COF design into a unified learning framework. By jointly modeling complementary motion phases and directly learning from optical flow inputs, FMANet achieves state-of-the-art performance across multiple benchmarks. Specifically, it establishes new SOTA results on SAMM (84.56\% accuracy, 0.810 UF1, 0.799 UAR in the 5-class setting), remains highly competitive on CASME-II (87.30\% accuracy, 0.807 UF1, 0.805 UAR in the 3-class setting), and demonstrates robust generalization on MMEW under multiple label configurations (3-, 5-, 6-, and 7-class).

This progression, from MM-COF as a strong handcrafted representation to FMANet as an integrated end-to-end architecture, demonstrates the effectiveness of phase-aware motion modeling for micro-expression recognition. Future work may explore more advanced attention mechanisms and domain generalization strategies to further enhance the robustness and cross-dataset applicability of FMANet.

\bibliographystyle{IEEEtran}
\bibliography{references}

\end{document}